\documentclass{article}
\usepackage{spconf,amsmath,graphicx}
\usepackage{hyperref}
\usepackage{booktabs}
\usepackage{amssymb}
\usepackage{multirow}
\usepackage{xcolor}
\usepackage{balance}
\usepackage{pifont}
\usepackage{amsmath}

\usepackage{enumitem}
\setlist{nosep, leftmargin=14pt}

\title{
MMA-Former: Multi-Window Mixture-of-Head Attention Transformer \\
for Adaptive PNI Prediction in 3D MRI
}

\name{%
\begin{tabular}{@{}c@{}}
Youngung Han$^{1,2}$, Induk Um$^{3}$, Kyeonghun Kim$^{2}$,
Junga Kim$^{1}$, Hyunsu Go$^{1}$, Jaewon Jung$^{1}$, \\
Woo Kyoung Jeong$^{4}$, Won Jae Lee$^{5}$,
Pa Hong$^{5}$, Ken Ying-Kai Liao$^{6}$,
Hyuk-Jae Lee$^{1}$, Nam-Joon Kim$^{1,\dagger}$
\end{tabular}
}

\address{
$^{1}$Seoul National University, Seoul, Republic of Korea \\
$^{2}$OUTTA, Seoul, Republic of Korea \\
$^{3}$Chung-Ang University, Seoul, Republic of Korea \\
$^{4}$Samsung Medical Center, Sungkyunkwan University School of Medicine, Seoul, Republic of Korea \\
$^{5}$Samsung Changwon Hospital, Changwon, Republic of Korea \\
$^{6}$NVIDIA AI Technology Center, Taipei, Taiwan \\[0.3em]
$^{\dagger}$Corresponding author: \texttt{knj01@snu.ac.kr}
}

\begin{document}
\ninept

\maketitle

\begin{abstract}
Perineural invasion (PNI) is a critical prognostic factor in cholangiocarcinoma. Non-invasive prediction from 3D MRI is challenging, demanding models that efficiently capture both fine-grained details and global context. We propose the Multi-window Mixture-of-Head Attention Transformer (MMA-Former), a novel end-to-end 3D architecture featuring a Coarse-Fine Transformer (CFT) structure for parallel multi-scale feature extraction. We advance this structure by integrating a novel Window-Specific Mixture-of-Head attention (WS-MoH) mechanism. Unlike standard Multi-Head Self Attention (MSA), WS-MoH generates a representation for each 3D window and dynamically routes the entire window to specialized or common attention heads. This enables spatially adaptive feature extraction tailored to the local context of each window, enhancing specialization and reducing redundancy without increasing parameters. Evaluated on a retrospective dataset of 168 T1-weighted MRI scans, MMA-Former achieved an AUC of 0.752, outperforming other 3D architectures, including the best CNN (AUC of 0.708) and Transformer baselines (AUC of 0.681).
\end{abstract}

\section{Introduction}
\label{sec:intro}

Perineural invasion (PNI), the insidious infiltration of cancer cells along nerve sheaths, is a critical route of metastasis in cholangiocarcinoma. Its presence significantly escalates the risk of recurrence, correlates with poor survival, and dictates surgical planning ~\cite{zou2023perineural, zhang2020perineural, liu2024noninvasive}. Accurate preoperative identification of PNI is therefore paramount for personalized treatment.

\begin{figure}[h]
  \centering
  \centerline{\includegraphics[width=\columnwidth]{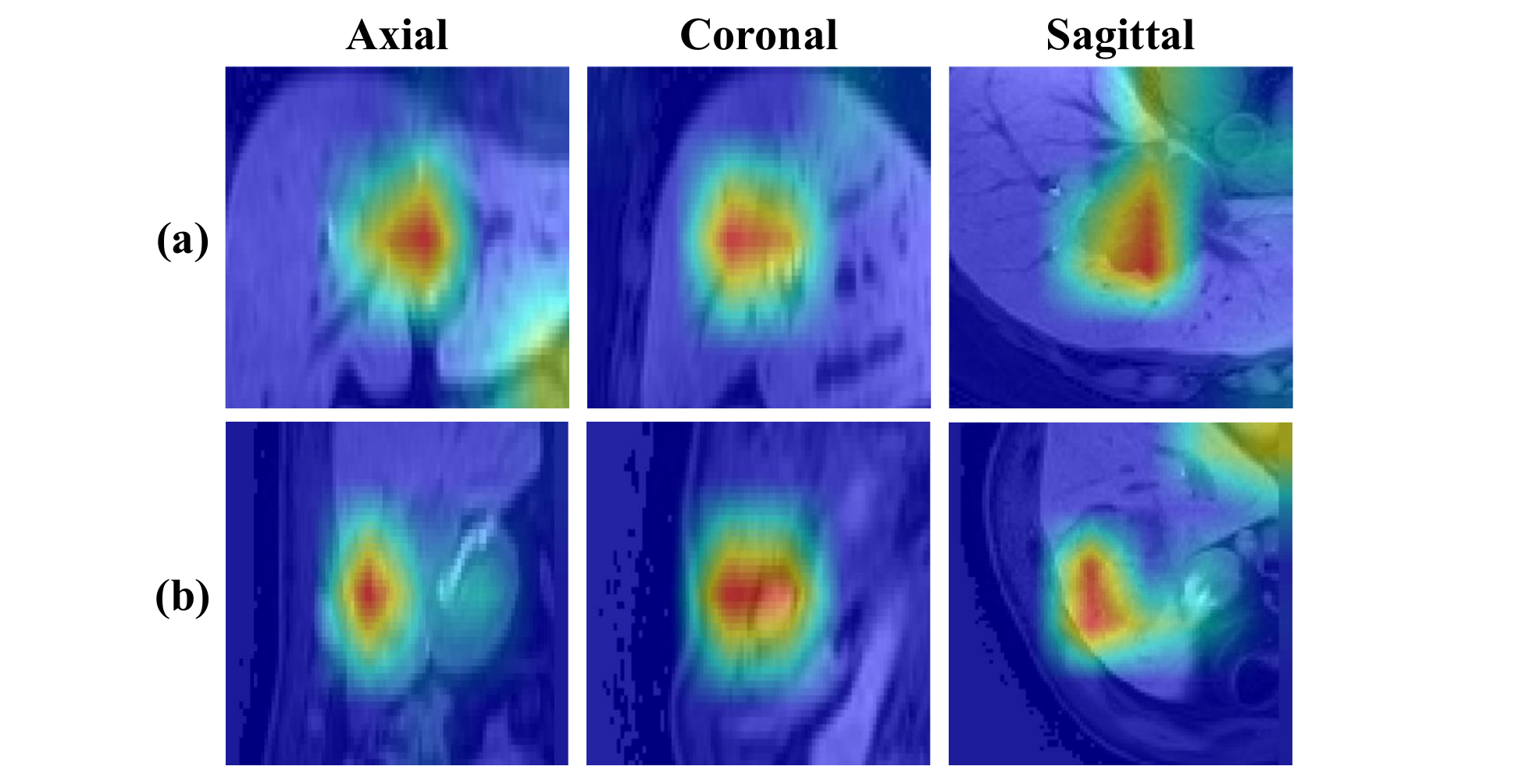}}
  \caption{Grad-CAM visualizations, highlighting critical regions at the tumor interface. (a) PNI-positive and (b) PNI-negative cases localized by MMA-Former.}
  \label{fig:gradcam}
\end{figure}

Despite its clinical urgency, non-invasive PNI prediction remains challenging due to subtle MRI features \cite{doran2024perineural, qi2025mri}. Furthermore, most PNI studies are hampered by small cohorts, often numbering in the low hundreds \cite{zhang2020perineural, zhan2022ct}, which frequently leads to class imbalance.

Methodologically, existing approaches often rely on radiomic features \cite{huang2021feasibility}, which may fail to capture complex 3D spatial patterns. While end-to-end deep learning offers potential, standard architectures struggle. CNNs \cite{he2016deep, huang2017densely} are limited in modeling long-range dependencies, while Vision Transformers (ViT) \cite{dosovitskiy2020image, vaswani2017attention} and their 3D adaptations \cite{hatamizadeh2022unetr} incur prohibitive computational costs in 3D MRI.

Hierarchical approaches like Swin Transformer \cite{liu2021swin} improve efficiency by computing attention within local windows. Building on this, architectures incorporating parallel multi-scale window processing, such as MViT \cite{fan2021multiscale} or Focal Transformers \cite{yang2021focal}, have emerged to explicitly capture features at different scales simultaneously. However, the uniform processing of standard Multi-Head Self Attention (MSA) in these methods limits adaptive feature selection and causes attention head redundancy.\cite{jin2024moh, voita2019analyzing, michel2019sixteen}.


The Mixture-of-Head attention (MoH) mechanism \cite{jin2024moh}, inspired by Mixture-of-Experts (MoE) principles \cite{shazeer2017outrageously, fedus2022switch}, addresses this redundancy by treating attention heads as experts. MoH employs a router to dynamically select a subset of specialized heads for each input. This enhances specialization and efficiency. However, MoH typically operates at the token level, which remains computationally intensive for 3D data.

\begin{figure*}[t!]
  \centering
  \centerline{\includegraphics[width=0.95\textwidth]{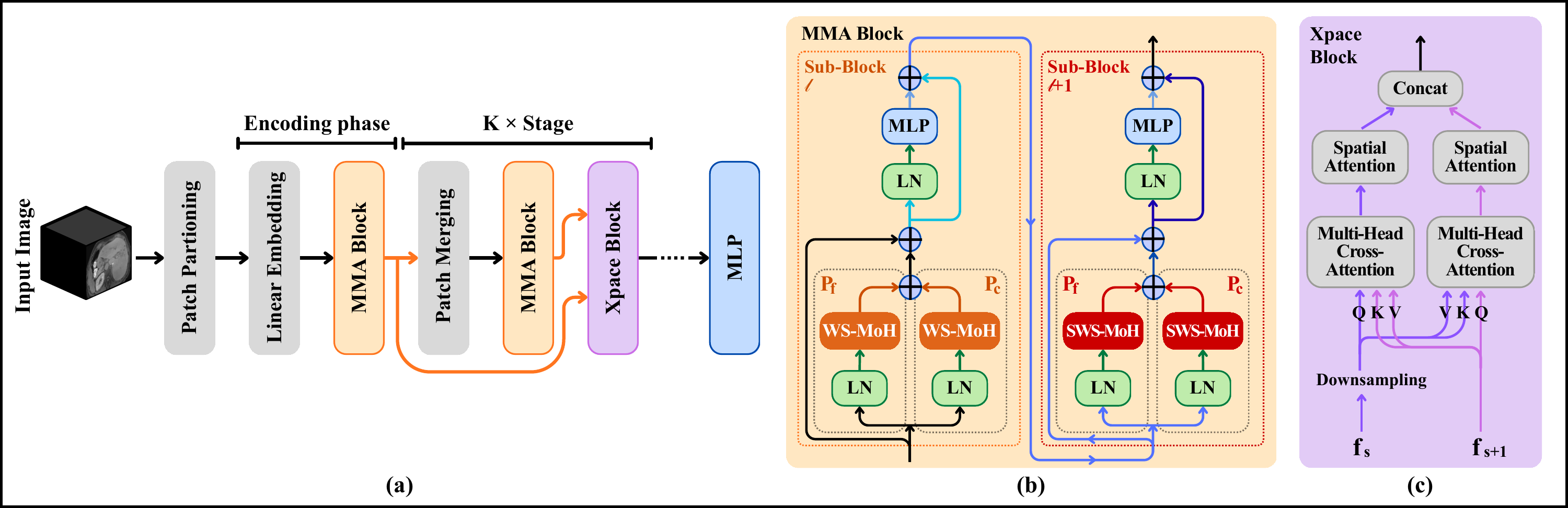}}
  \caption{Overview of the MMA-Former architecture. (a) The framework utilizes a hierarchical structure incorporating Multi-window Mixture-of-Head Attention (MMA) blocks and Cross-Spatial Attention (Xpace) blocks. (b) The MMA block implements the CFT structure across consecutive blocks (block $l$ and $l+1$). It utilizes parallel Coarse ($P_c$) and Fine ($P_f$) pathways with different window sizes, employing Window-Specific MoH (WS-MoH) and Shifted Window-Specific MoH (SWS-MoH). (c) The Xpace block for hierachical feature fusion through Multi-Head Cross-Attention and Spatial Attention.}
  \label{fig:overview}
\end{figure*}

We propose the Multi-window Mixture-of-Head Attention Transformer (MMA-Former), an adaptive 3D architecture that synergizes multi-scale processing with a novel adaptation of MoH. We introduce the Coarse-Fine Transformer (CFT) structure to capture parallel multi-scale information. Crucially, we integrate Window-Specific MoH (WS-MoH). Instead of routing every token, WS-MoH dynamically routes the entire window to specialized heads. This enables spatially adaptive feature extraction where different regions leverage different attention heads efficiently. 

Our main contributions are these:
\begin{itemize}
    \item MMA-Former, a novel 3D end-to-end architecture utilizing a CFT structure for parallel multi-scale feature extraction for PNI prediction.
    \item WS-MoH, a novel approach applying MoH routing at the window level, enabling adaptive feature extraction based on the spatial context of the window.
    \item Xpace Block, an encoder utilizing hierarchical feature fusion between different stages.
\end{itemize}


\begin{figure*}[t]
  \centering
  \centerline{\includegraphics[width=0.95\textwidth]{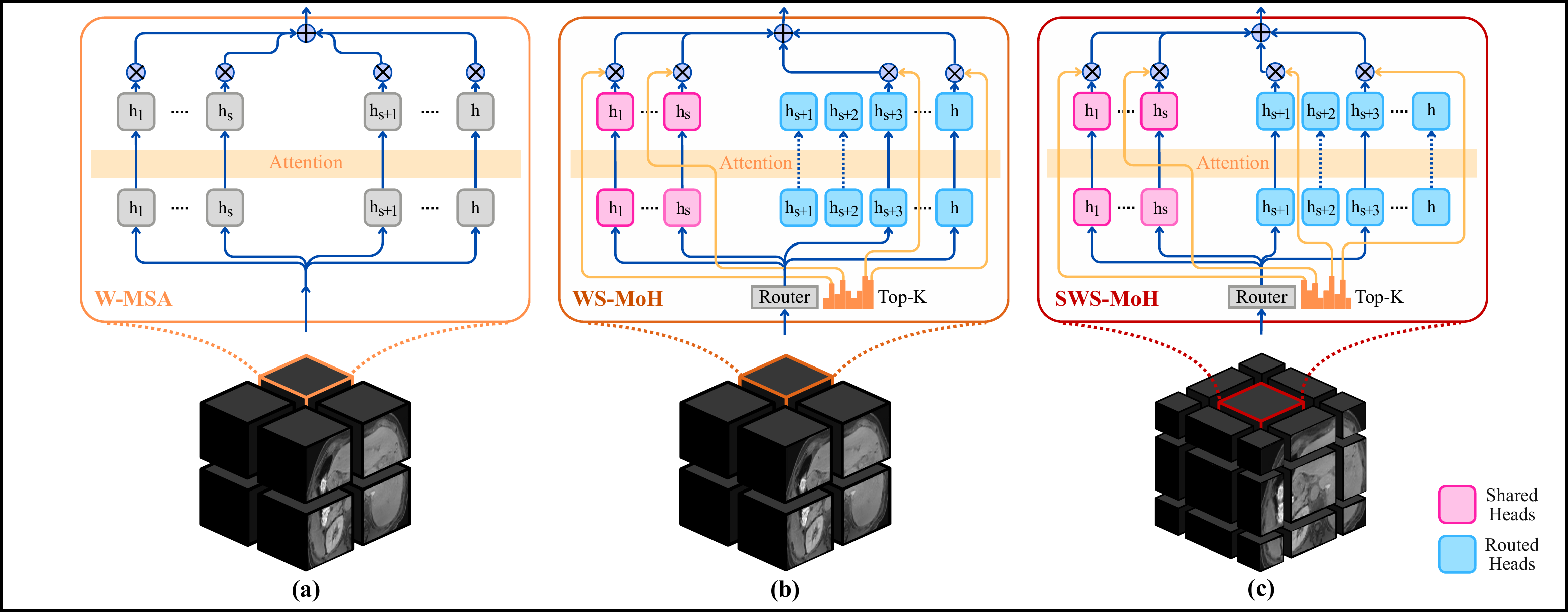}}
  \caption{Comparison of attention mechanisms. (a) W-MSA: All heads are activated uniformly for the window. (b) WS-MoH (Window-Specific MoH): A router utilizes window-representative features to select a Top-K subset of Routed Heads, which are combined with always-active Shared Heads. This allows different windows to utilize different head combinations. (c) SWS-MoH (Shifted Window-Specific MoH): WS-MoH applied to a shifted window configuration. As the window is shifted, the router dynamically activates different head combinations for each respective window.}
  \label{fig:architecture_details}
\end{figure*}

\section{Methodology}
\label{sec:method}

\subsection{Datasets and Preprocessing}
\label{ssec:data}

We utilized a retrospective dataset of anonymized T1-weighted, contrast-enhanced MR images (hepatobiliary phase) acquired from multiple MRI scanners at Samsung Medical Center over a decade. Images were provided in NIfTI format with ground truth annotations for the liver and tumor. After quality control, the final analysis cohort comprised 168 cholangiocarcinoma patients. The presence of PNI was confirmed by post-surgical histopathological examination, comprising 67 PNI-positive cases and 101 PNI-negative cases.

We employed a localization strategy, extracting cropped volumes of size $96\times96\times48$ centered on the tumor. The necessity of this localization is validated in Table~\ref{tab:ablation} (A).

\subsection{MMA-Former Architecture}
\label{ssec:architecture}

The MMA-Former, illustrated in Fig.~\ref{fig:overview} (a), employs a hierarchical design. Input 3D volumes undergo patch partitioning and linear embedding. The architecture consists of an encoding phase followed by $K$ stages. Each stage comprises patch merging layer and an MMA block for feature extraction. An Xpace block, shown in Fig.~\ref{fig:overview} (c), fuses hierarchical features across stages by integrating downsampled representations from the previous stage with features from the current stage. 

\subsection{The MMA Block: CFT with WS-MoH}
\label{ssec:mma_module}

\subsubsection{CFT Structure}
The Coarse-Fine Transformer (CFT) structure captures multi-scale information by processing features through parallel pathways with different window sizes.

Given an input feature map $z^{l-1}$ at sub-block $l$, we apply layer normalization (LN). The normalized features $z'$ are processed in parallel through two pathways. The Fine path, denoted as $P_f$, utilizes a small window size $W_F$ (e.g., $3\times3\times3$). The Coarse path, denoted as $P_c$, employs a larger window size $W_C$ (e.g., $6\times6\times6$). Both pathways use WS-MoH as the attention mechanism.
\begin{equation}
\label{eq:cft_parallel}
A_F = \text{WS-MoH}_{W_F}(z'), \quad A_C = \text{WS-MoH}_{W_C}(z')
\end{equation}
The outputs are fused via concatenation and linear projection (Fusion), then combined with the input via residual connections, followed by an MLP:
\begin{equation}
\label{eq:cft_aggregate}
A_{fused} = \text{Fusion}(A_F, A_C) + z^{l-1}
\end{equation}
\begin{equation}
\label{eq:cft_mlp}
z^{l} = \text{MLP}(\text{LN}(A_{fused})) + A_{fused}
\end{equation}
Subsequent blocks (Block $l+1$) employ shifted window partitioning (SWS-MoH) \cite{liu2021swin}
for cross-window communication (Fig.~\ref{fig:overview} (b)).

\subsubsection{WS-MoH}
We replace standard MSA with the adaptive Window-Specific MoH (WS-MoH), detailed in Fig.~\ref{fig:architecture_details}. MSA (Eq. \ref{eq:msa}) activates all $h$ heads uniformly, regardless of the input context.
\begin{equation}
\label{eq:msa}
\text{MSA}(X) = \sum_{i=1}^{h} H^i(X) W_O^i
\end{equation}

MoH \cite{jin2024moh} introduces a router to dynamically weight heads, denoted as $g_i$. We propose WS-MoH, which adapts this routing to the window level for efficiency in 3D data.

First, standard Q, K, V projections are computed for the input features within a window $X_W$.
\begin{equation}
\label{eq:qkv}
Q = X_W W_Q, \quad K = X_W W_K, \quad V = X_W W_V
\end{equation}

Crucially, instead of routing each token, we generate a window representative feature $X_{rep}$ by average pooling the features $X_W$.

The router uses this single representative feature $X_{rep}$ to determine the routing scores $g_i$ for the entire window. The final WS-MoH output is the weighted sum of the outputs from the selected heads:
\begin{equation}
\label{eq:w-moh}
\text{WS-MoH}(X_W) = \sum_{i=1}^{h} g_i(X_{rep}) H^i(X_W) W_O^i
\end{equation}

This enables spatially adaptive feature extraction, as different windows activate different combinations of heads based on their local context.

We employ the two-stage routing strategy \cite{jin2024moh}. Heads are divided into Shared Heads, denoted as $H_S$, which capture common knowledge and are always active, and Routed Heads, denoted as $H_R$, which handle specialized patterns.

A router network processes $X_{rep}$ to produce probabilities for the routed heads, and the Top-K heads are selected.
\begin{equation}
\label{eq:router_probs}
P_R = S(W_{r}X_{rep})
\end{equation}

The routing score $g_i$ (Eq. \ref{eq:routing}) is determined by balancing the contributions of $H_S$ and the Top-K selected $H_R$ via coefficients $\alpha_1, \alpha_2$, where $[\alpha_{1},\alpha_{2}] = S(W_{h}X_{rep})$ and $S(\cdot)$ denotes the Softmax function.
\begin{equation}
\label{eq:routing}
g_{i}(X_{rep})=\begin{cases}
\alpha_{1} S(W_{s}X_{rep})_{i}, & \text{if } i \in H_S \\
\alpha_{2} (P_R)_{i}, & \text{if } i \in \text{Top-K}(H_R) \\
0, & \text{otherwise}.
\end{cases}
\end{equation}

\begin{table}[h]
\centering
\caption{AUC comparison of PNI classifiers on cropped 3D images (5-fold CV).}
\label{tab:performance}
\resizebox{\columnwidth}{!}{%
\begin{tabular}{@{}llc@{}}
\toprule
\textbf{Category} & \textbf{Model (3D)} & \textbf{Mean AUC} \\ \midrule
CNN & ResNet \cite{he2016deep} & 0.708 \\
    & DenseNet \cite{huang2017densely} & 0.688 \\
    & EfficientNet\cite{tan2019efficientnet} & 0.673 \\ \midrule
Transformer & Swin Transformer \cite{liu2021swin} & 0.681 \\
            & \textbf{MMA-Former (CFT + WS-MoH)} & \textbf{0.752} \\ \bottomrule
\end{tabular}%
}
\end{table}

\subsubsection{Load Balance Loss}
\label{sssec:load_balance}
In mixture models utilizing routing, training often leads to an imbalance where a few heads process the majority of inputs, leaving others undertrained \cite{shazeer2017outrageously}. To mitigate this in WS-MoH and ensure effective utilization of all Routed Heads for robust PNI prediction, we incorporate a load balance loss $\mathcal{L}_{LB}$. This loss encourages an even distribution of windows across the Routed Heads.
\begin{equation}
\label{eq:load_balance}
\mathcal{L}_{LB} = \sum_{i \in H_R} P_i \cdot f_i
\end{equation}
where $P_i$ is the average routing probability for head $i$, and $f_i$ is the fraction of windows that selected head $i$.

\subsection{Xpace Block }
\label{sssec:xpace_block}
The Cross-Spatial Attention (Xpace) block fuses features across encoder stages. First, it downsamples previous-stage features. The fusion mechanism operates through bidirectional cross-attention between downsampled and current-stage features. Spatial attention is applied to highlight important regions in both representations. The resulting features are then concatenated along the channel dimension and forwarded to the next stage.

\begin{table*}[t]
\centering
\caption{Comprehensive ablation study of MMA-Former components and configurations on the PNI dataset.}
\label{tab:ablation}
\resizebox{\textwidth}{!}{
\begin{tabular}{@{}lllc@{}}
\toprule
\textbf{Category} & \textbf{Configuration} & \textbf{Description} & \textbf{AUC} \\ \midrule
\textbf{Baseline} & MMA-Former (Full Model) & Cropped Input, CFT (2 paths), WS-MoH (S=2, Top-K 75\%), Xpace & \textbf{0.752} \\ \midrule
\textit{(A) Input Strategy} & Uncropped Input & MMA-Former trained on whole MRI volumes & 0.705 \\
& Cropped Input (Default) & MMA-Former trained on localized tumor volumes ($96\times96\times48$) & \textbf{0.752} \\ \midrule
\textit{(B) Component Ablation} & w/o WS-MoH (CFT+MSA) & WS-MoH replaced by standard MSA (All heads active) & 0.731 \\
& w/o CFT Structure & Standard Transformer blocks + WS-MoH (Single window path) & 0.690 \\
& w/o Xpace Block & Hierarchical features not explicitly fused via Xpace & 0.709 \\ \midrule
\textit{(C) CFT: Window Paths} & Fine Path Only & Single path utilizing only the small window size & 0.704 \\
& Coarse Path Only & Single path utilizing only the large window size  & 0.718 \\
& Fine + Coarse (Default) & Parallel Fine  + Coarse paths & \textbf{0.752} \\ \midrule
\textit{(D) WS-MoH: Shared Heads (S)} & S=0 & No Shared Heads; Only Routed heads active (Top-K 75\%) & 0.748 \\
(Top-K ratio fixed at 75\%) & S=1 & 1 Shared head + Routed heads (Top-K 75\%) & 0.737 \\
& S=2 (Default) & 2 Shared heads + Routed heads (Top-K 75\%) & \textbf{0.752} \\
& S=All & All Shared; No routing (Equivalent to MSA) & 0.731 \\ \midrule
\textit{(E) WS-MoH: Top-K Ratio} & 50\% & S=2, Top-K 50\% of Routed heads active & 0.743 \\
(S fixed at 2) & 75\% (Default) & S=2, Top-K 75\% of Routed heads active & \textbf{0.752} \\
& 90\% & S=2, Top-K 90\% of Routed heads active & 0.733 \\
\bottomrule
\end{tabular}
}
\end{table*}

\section{Experiments and Results}
\label{sec:experiments}
\subsection{Experimental Setup}
All experiments were performed on NVIDIA A100 GPUs. We employed stratified 5-fold cross-validation across the 168 cases. Models were trained using the AdamW optimizer \cite{loshchilov2017decoupled} with a learning rate of $8\times10^{-5}$ and a batch size of 4. The total loss function combines the Weighted BCE Loss, denoted as $\mathcal{L}_{WCE}$, utilizing a positive weight of 1.5 to address class imbalance, and the load balance loss $\mathcal{L}_{LB}$, weighted by a factor $\beta=0.01$.
\begin{equation}
\label{eq:total_loss}
\mathcal{L}_{Total} = \mathcal{L}_{WCE} + \beta\mathcal{L}_{LB}
\end{equation}

\subsection{PNI Prediction Performance}

Table~\ref{tab:performance} summarizes the performance comparison. MMA-Former achieved the highest mean AUC of 0.752. It significantly outperformed the best-performing CNN, ResNet (which achieved an AUC of 0.708), and the standard 3D Swin Transformer (which achieved an AUC of 0.681).

\subsection{Ablation Study}
\label{ssec:ablation}

We conducted ablation studies to validate the design choices of MMA-Former (Table~\ref{tab:ablation}). The default configuration uses 2 Shared heads (S=2) and a Top-K ratio of 75\%.

\textbf{Impact of Input Strategy (A):} Training MMA-Former on uncropped images decreased the AUC by 0.047. This confirms that localization to the tumor region is crucial for detecting the subtle features of PNI, validating our preprocessing strategy.

\textbf{Impact of Core Components (B):} Replacing WS-MoH with standard MSA decreased the AUC by 0.021. This highlights that the adaptive, window-specific feature extraction provided by WS-MoH contributes to performance. Removing the CFT structure also reduced performance, resulting in an AUC of 0.690. Additionally, removing the Xpace Block resulted in degraded performance, lowering the AUC to 0.709.

\textbf{Impact of CFT Configuration (C):} The parallel configuration achieved the best performance, outperforming both the Fine path only (AUC of 0.704) and the Coarse path only (AUC of 0.718). This validates the CFT structure, confirming that synergizing local and global contexts outperforms single-scale approaches.

\textbf{Impact of WS-MoH Shared Heads (D):} The configuration with 2 Shared heads provided the optimal balance. Relying solely on Routed heads failed to capture common features effectively, while using only Shared heads lacked adaptive specialization.

\textbf{Impact of WS-MoH Top-K Ratio (E):} Activating 75\% of the routed heads yielded the best result. Using fewer heads reduced capacity, while using more likely reintroduced redundancy.

\subsection{Qualitative Results and Visualization}

We utilized 3D Grad-CAM \cite{selvaraju2017grad} to visualize the regions most influential to the model's predictions (Fig.~\ref{fig:gradcam}). The visualizations confirm that MMA-Former focuses precisely on the tumor and the immediate peritumoral interface, the areas most critical for identifying PNI. 

\section{Discussion and Conclusion}

The MMA-Former introduces a novel approach to 3D medical image analysis by enabling spatially adaptive feature extraction within a hierarchical transformer framework. The effectiveness of the approach is strongly supported by the necessity of input localization.

The core innovation is the WS-MoH within the CFT framework. The CFT structure provides multi-scale context. WS-MoH significantly outperforms CFT+MSA via window-level adaptation. Dynamically routing window representations to specialized heads captures nuanced local context. This contrasts with MSA's uniform processing and avoids the computational burden of token-level MoH.

The primary limitation is the single-institution nature of the dataset, comprising 168 cases. Although the cohort size is comparable to related PNI studies, external validation on multi-center datasets is required to ensure generalizability.

In conclusion, we proposed the MMA-Former, an adaptive 3D Transformer for PNI prediction. By synergizing the CFT structure with a novel WS-MoH, MMA-Former effectively captures multi-scale context while enabling efficient, spatially adaptive feature extraction. Our experiments demonstrate the superiority of MMA-Former (AUC of 0.752) over existing 3D backbones (best AUC of 0.708), highlighting the benefit of window-level adaptive attention for complex 3D medical imaging tasks.

\section{Acknowledgments}
This work was supported by the Institute of Information \& Communications Technology Planning \& Evaluation (IITP), funded by the
Korea government (MSIT), under the Artificial Intelligence Semiconductor Support Program to nurture the best talents (IITP-2023-
RS-2023-00256081) and the grant for the Development of an AI
Deep Learning Processor and Module for a 2,000 TFLOPS Server
(No. 2020-0-01305)

{\small
\bibliographystyle{IEEEtran}
\bibliography{refs}
}

\end{document}